\documentclass[letterpaper]{article} 
\usepackage{aaai2026}  
\usepackage{times}  
\usepackage{helvet}  
\usepackage{courier}  
\usepackage[hyphens]{url}  
\usepackage{graphicx} 
\usepackage{natbib}  
\usepackage{caption} 
\usepackage{amsmath, amssymb}
\usepackage{comment}
\usepackage{algorithm}
\usepackage{algorithmic}
\usepackage{booktabs}
\usepackage{multirow}
\usepackage{subfig}
\usepackage{tablefootnote}
\usepackage{xspace}

\usepackage[most]{tcolorbox}
\usepackage{lipsum} 
\usepackage{tcolorbox}
\tcbuselibrary{raster}
\usepackage{booktabs}
\usepackage{newfloat}
\usepackage{listings}
\usepackage{xcolor} 
\DeclareCaptionStyle{ruled}{labelfont=normalfont,labelsep=colon,strut=off} 
\floatstyle{ruled}
\newfloat{listing}{tb}{lst}{}
\floatname{listing}{Listing}
\title{Perturb Your Data: Paraphrase-Guided Training Data Watermarking}
\author{
    Pranav Shetty,
    Mirazul Haque,\\
    Petr Babkin,
    Zhiqiang Ma,
    Xiaomo Liu,
    Manuela Veloso
}
\affiliations{
    JPMorgan AI Research\\

    \{first.last\}@jpmchase.com
}

\newcommand{\todo}[1]{\textbf{todo:#1}}

\newcommand{\method}{\textsc{SPECTRA}}
\newcommand{\minkpp}{Min-K\%++ }

\begin{document}

\maketitle

\begin{abstract}

Training data detection is critical for enforcing copyright and data licensing, as Large Language Models (LLM) are trained on massive text corpora scraped from the internet. We present SPECTRA, a watermarking approach that makes training data reliably detectable even when it comprises less than 0.001\% of the training corpus. 
SPECTRA works by paraphrasing text using an LLM and assigning a score based on how likely each paraphrase is, according to a separate scoring model. A paraphrase is chosen so that its score closely matches that of the original text, to avoid introducing any distribution shifts. To test whether a suspect model has been trained on the watermarked data, we compare its token probabilities against those of the scoring model.
We demonstrate that SPECTRA achieves a consistent p-value gap of over nine orders of magnitude when detecting data used for training versus data not used for training, which is greater than all baselines tested. SPECTRA equips data owners with a scalable, deploy‑before‑release watermark that survives even large‑scale LLM training.

\end{abstract}

\section{Introduction}

Contemporary large language models (LLMs) utilize extensive datasets sourced from the internet for pretraining, which lead to their general-purpose abilities, but may include content scraped without permission. Although these large corpora are necessary for the emergent abilities of LLMs, this practice raises several ethical and legal concerns. Firstly, the utilization of copyrighted material in the training process may violate its licensing terms. Secondly, the incorporation of widely recognized benchmarking datasets during pretraining could compromise the integrity of model evaluation. Many open-weight models are released without disclosing their training data, which leaves model end-users vulnerable to liability for damages and can hinder the adoption of open-source models.

Several recent lawsuits have focused on the unauthorized use of pay-walled data that was used for training models \cite{reuters_nytimes_openai_2023, brittain_authors_sue_anthropic_2024}. As more and more content is consumed online through various intermediate LLMs like ChatGPT, it becomes vital to ensure that content creators are appropriately incentivized to produce new content. Otherwise, the data collection practices of most LLM providers may lead to an ``extractive dead end'' \cite{rosenblat2025beyond}. Consequently, there is an important need to detect unauthorized use of data for training.

Many recent techniques have been developed to detect training data \cite{yeom2018privacy, zhang2025mink, zhang-etal-2024-pretraining}. Membership Inference Attack (MIA) techniques build on the idea that training leads to changes in the probabilities output by the model, which is measured by comparing these probabilities (MIA scores) for training data (also called member data) against a held-out or non-member dataset from the same domain that was not used for training. A content creator must provide the suspect data and held-out data, which can be used to test a target model. However, MIA techniques are sensitive to small distribution shifts between the suspect and held-out data that can lead to spurious performance \cite{duan2024membership, mainidi2024, zhaounlocking}.

\begin{figure}
    \centering
    \includegraphics[width=1.02\linewidth]{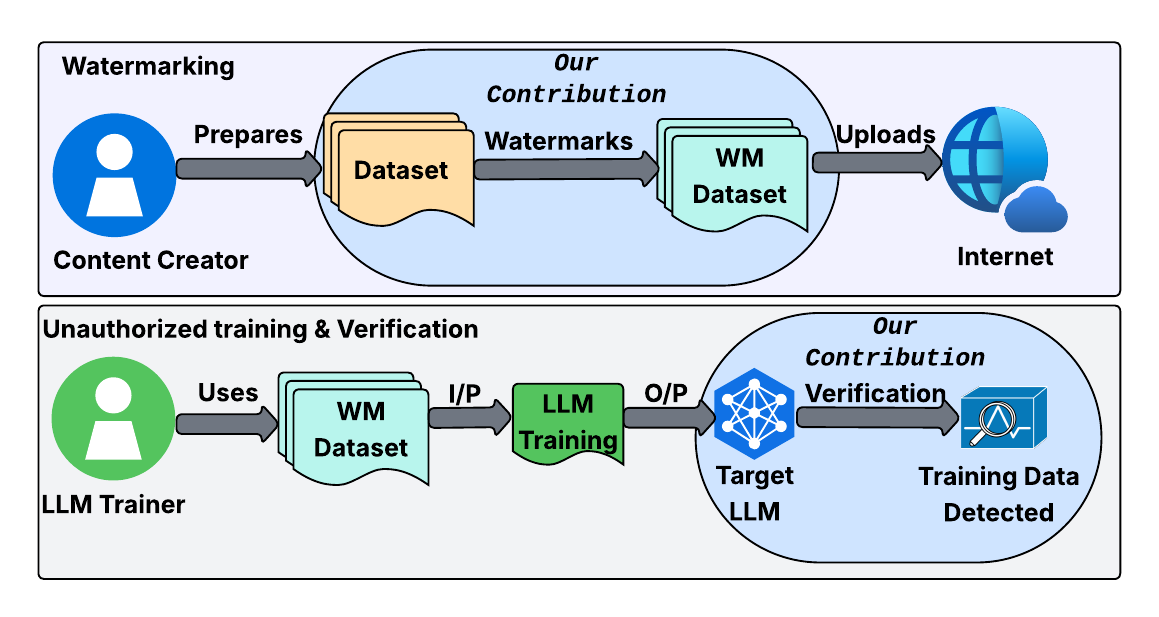}
    \caption{Overview of our problem setting. I/P and O/P refer to input and output, respectively.}
    \label{fig:overview}
\end{figure}


Given the limitations of standard MIA techniques, it is necessary to find alternative ways through which content creators can protect their content. STAMP \cite{rastogi2025stampcontentprovingdataset} is a recent effort in this direction that watermarks text using the KGW scheme \cite{kirchenbauer2023reliability} multiple times with different keys, with one version made public and the rest kept private by the content creator. If their data is used for training, content creators can compare the perplexity of the public rephrases against the private rephrases to establish membership (included in the training data). In contrast to MIA, which detects the membership of each document, STAMP performs inference over the entire dataset, enabling it to be robust to noise in individual examples. STAMP requires storing a large number of private rephrases for each document. This method also requires access to the decoding layers of an LLM to generate watermarked text, which may be inaccessible due to the large amount of GPU resources required. The p-values between member and non-member datasets are at most three orders of magnitude apart in STAMP, which we argue is not sufficient for such a task with significant legal implications.


To address these limitations, we introduce \textbf{S}core sampled re\textbf{P}hrasing to det\textbf{EC}t \textbf{TRA}ining data (\method) (Figure \ref{fig:overview}). We assume that a content creator creates textual content and, before making it publicly available, watermarks a portion of it using \method. Such content might be openly accessible, placed behind a paywall, or protected by licenses explicitly prohibiting web scraping for model training. If the content creator suspects a particular model of unauthorized use of their data, they or an authorized intermediary can conduct a statistical test to confirm whether the watermarked data was indeed part of that model’s training data.

The \method{} approach can be divided into two phases: watermarking and verification. 
During \textit{watermarking}, \method{} generates several paraphrases using an LLM without needing to modify its decoding process. A score that provides a membership signal is computed over the paraphrases using a scoring model that has not been trained on this data. Specifically, we compute the \minkpp~\citep{zhang2025mink} score, which is the normalized log probability averaged over the lowest \(K\%\) tokens (highest surprisal tokens) in a document. Our key insight is that existing MIA scores are designed to measure changes in the loss surface between a model before and after it is trained on specific data, but in practice, we seldom have access to the model before training. We employ the scoring model as a proxy for this initial (pre-trained) state, allowing us to detect changes attributable to training reliably. To avoid introducing systematic bias, paraphrases are carefully selected using a sampling strategy that we designed, such that their \minkpp scores remain close to the original document's score. This constraint ensures that the watermarking procedure itself does not introduce any false-positive membership signals. In the \textit{verification} phase, we construct a statistical test in which the scores from the \minkpp scores from the scoring model are compared against the scores from the suspected model to obtain a p-value. Unlike other MIA methods, \method{} does not require a non-member dataset. We find that \method{} yields a statistically significant result for identifying membership in all datasets used for training, without yielding false positives. \method{} will equip content creators with the tools to enforce their intended data-use policies.

Our contributions are as follows:

\begin{enumerate}
    \item We show that \method\ can be used to watermark pre-training data and that the watermark can be measured after continued pre-training with 5 billion tokens. Each of our datasets constitute less than 0.001\% of the training corpus.
    \item We design a strategy to sample paraphrases that outperforms other sampling approaches, such as random sampling or selecting paraphrases with the maximum \minkpp{} score. 
    \item We benchmark against existing methods for detecting training data and find that \method{} is the only one that yields a statistically significant result to identify membership for all datasets tested. Moreover, \method{} gave the largest change in p-values between non-members and members, yielding a consistent difference of at least nine orders of magnitude across datasets using $500$ samples each.
\end{enumerate}

\section{Related Work}

\textbf{Membership Inference Attack (MIA)}.
In the context of LLMs, recent studies have proposed several Membership Inference Attack methods that use scores derived from the log probabilities output by the model to differentiate member data (used during training) from non-member data (not used during training). The datasets employed to evaluate these methods, such as WikiMIA \cite{shi2023detecting} and PatentMIA \cite{zhang-etal-2024-pretraining}, were constructed by collecting data published before (member) and after (non-member) the LLM's knowledge cutoff date. It was later observed that these MIA techniques primarily detected temporal artifacts. Subsequent work showed that when member and non-member datasets were sampled homogeneously—thereby removing temporal signals—all tested MIA methods performed no better than random classifiers \cite{duan2024membership, mainidi2024, das2025blind}.


\noindent\textbf{Data Watermarks}.
Data watermarks are modifications made to a text to make it more detectable when used to train a model. \citet{wei-etal-2024-proving} insert hash strings into a model and show that models trained on such hashes occurring multiple times in the dataset memorize the hash. Other works have proposed backdoor attacks that insert carefully picked tokens into the text, which can be detected post-training by using a secret prompt \cite{bouaziz2025winter, yan2022textual}. The assumption underlying these works is that LLM trainers collect large quantities of data from the internet and are likely to collect data containing such watermarks. The challenge is that such watermarks affect the meaning and readability of the text and are thus not suitable when the text is meant for human readers. More recently, STAMP addresses this issue by rephrasing text from instruction tuning benchmarks using the Llama3-70b model and watermarking the text using the KGW scheme.

\noindent\textbf{Dataset Inference}. 
In contrast to MIA, \citet{mainidi2024} propose Dataset Inference (DI) where the goal is not to obtain accurate labels over every document in the dataset but to obtain a measure of confidence over the entire dataset being tested. As LLM trainers tend to scrape each source comprehensively, it is likely that related documents from a single source will all be used for training. Using multiple documents also enhances the signal available for detecting membership and makes the inference less susceptible to noise due to outliers. Our work takes inspiration from this.



\section{Problem Setup}

Consider a content creator who possesses a dataset \( D \), which they wish to make available online. \( D \) could consist of articles from sources such as news providers or academic publishers. The creator, however, wishes to prevent unauthorized use of this dataset for training LLMs.

To enable detection of unauthorized use, the creator applies a watermarking procedure \(W\) to transform the original dataset \( D \) into a modified dataset \( D' = W(D) \). 
The creator retains \( D \) and only publishes \( D' \).

After publication, the content creator may seek to test whether a particular target model \( M \) has been trained on the watermarked dataset \( D' \). We assume a grey-box setting, where the model \( M \) is queried and provides log probabilities over tokens, but the model weights and architecture are not necessarily known. This scenario is typical of open-weight models, some of which are used commercially. For closed-source models, testing may be facilitated through a neutral third-party arbiter with grey-box access (say, a court-appointed arbiter). Given \(D\), \(D'\) and grey-box access to $M$, the content creator or arbiter applies a statistical test \( T \) to determine if the model \( M \) was trained on \( D' \). Note that this procedure detects membership of the entire dataset and not each document, which can be noisier.



The research question we address is:
How can we optimally design the watermarking procedure \( W \) and the statistical test \( T \) such that:
\begin{enumerate}
    \item If the model \( M \) is indeed trained on the watermarked dataset \( D' \), then \( T \) reliably identifies this fact with high confidence.
    \item Conversely, if \( M \) is \emph{not} trained on \( D' \), then \( T \) does not yield false-positive outcomes.
\end{enumerate}

In the sections that follow, we propose a method to achieve these objectives and empirically validate its efficacy.

\subsection{Training Data Detection Signals}

Some common scores in the literature that are computed to determine membership in the training data of a model are described below. In our scenario, \(M\) is an autoregressive language model that generates a probability distribution for each subsequent token, denoted as \(P(x_t \mid x_{<t}; M)\).

\begin{enumerate}
    \item \textbf{Loss} \cite{yeom2018privacy}: This method relies on measuring the loss of a given target sequence \(x\) under the model \(M\). The membership inference score is directly defined as: $f(x; M) = L(x; M)$ where \(L(x;M)\) denotes the negative log-likelihood (loss) of the target sequence according to the model \(M\).

    \item \textbf{Min-K\%} \cite{shi2023detecting}: This method focuses specifically on the \(K\%\) of tokens that have the lowest likelihood under the model \(M\). The membership inference score is computed as the average log probability over these tokens.
    \item \textbf{\minkpp{}} \citep{zhang2025mink} \minkpp enhances the original Min-K\% score by incorporating normalization relative to the mean and variance of token log probabilities.
\end{enumerate}

We focus on \minkpp{} in this work, which we describe in greater detail next.


\subsection{\minkpp{} score}

Formally, given an autoregressive model \(M\) and a token sequence \( x = (x_1, x_2, \dots, x_n) \), define the token-level normalized log probability as:

\[
z(x_t; M) = \frac{\log P(x_t \mid x_{<t}; M) - \mu_{x_{<t}}}{\sigma_{x_{<t}}}
\]
where
\begin{align*}
\mu_{x_{<t}} &= \mathbb{E}_{z \sim P(\cdot \mid x_{<t}; M)} [\log P(z \mid x_{<t}; M)],\\
\sigma_{x_{<t}} &= \sqrt{\mathbb{E}_{z \sim P(\cdot \mid x_{<t}; M)} \left[(\log P(z \mid x_{<t}; M) - \mu_{x_{<t}})^2\right]}.
\end{align*}
Here, \(\mu_{x_{<t}}\) represents the expectation of the log probability distribution for the next token given the prefix \(x_{<t}\), and \(\sigma_{x_{<t}}\) denotes the corresponding standard deviation. 

The \minkpp{} score for a sequence \(x\) is defined as the average of the normalized log probabilities \(z(x_t; M)\) over the \(K\%\) of tokens in the sequence with the lowest values (indicating highest surprisal):

\[
f_{\text{Min-K\%++}}(x; M) = \frac{1}{|\text{min-K}(x)|}\sum_{x_t \in \text{min-K}(x)} z(x_t; M).
\]

This normalization allows \minkpp to better distinguish sequences that are part of the training data from those that are not, by highlighting the relative surprisal of the most unlikely tokens while making it robust to absolute probability shifts across tokens.
Critically, prior work \cite{zhang2025mink} has shown that the \minkpp score theoretically corresponds to measuring the negative trace of the Hessian of the log-likelihood \(log\; P(x_t \mid x_{<t})\). Intuitively, training via maximum-likelihood directly reduces the curvature (Hessian trace) of the loss landscape at training examples, thereby causing their corresponding \minkpp scores to increase. We evaluate the performance of various MIA scores on three datasets (Table \ref{tab:prelim_results}) that were not used during the pretraining of Pythia models. When the datasets, each containing 500 samples (at most 512 tokens each), are used for training with an additional 500 million text tokens, \minkpp has the best performance among all methods. However, as reported by prior work \cite{duan2024membership}, the effectiveness of \minkpp diminishes significantly at larger scales of training (5 billion tokens), dropping to performance indistinguishable from random. Based on these observations, we hypothesize that while \minkpp inherently captures strong training signals, its effectiveness at large scales of training may be hindered by distributional homogeneity when measured against non-member data, and that changing the reference point for measurement would enable us to capture a stronger training signal.



\begin{table}[htbp]
\centering
\begin{tabular}{@{}l|lccc@{}}
\toprule
& Metrics & \textbf{Wiki} & \textbf{HN} & \textbf{PubMed} \\
\midrule
\multirow{4}{*}{\parbox{2cm}{\centering\textbf{Datasets\\+500 million\\tokens}}} 
& Loss & 0.71 & 0.73 & 0.63 \\
& DC-PDD\tablefootnote{See appendix for more details} & 0.77 & 0.79 & 0.64 \\
& Min-k \% & 0.76 & 0.79 & 0.65 \\
& \minkpp & \textbf{0.85} & \textbf{0.84} & \textbf{0.72} \\
\midrule
\multirow{4}{*}{\parbox{2cm}{\centering\textbf{Datasets\\+5 billion\\tokens}}} 
& Loss & 0.55 & 0.54 & 0.52 \\
& DC-PDD & 0.55 & 0.52 & 0.50 \\
& Min-k \% & 0.56 & 0.55 & 0.52 \\
& \minkpp & 0.55 & 0.55 & 0.51 \\
\bottomrule
\end{tabular}
\caption{ROC-AUC of classifying training data used for continued pretraining of a Pythia 410m model against a held-out dataset from the same domain. The ROC-AUC is computed using the MIA scores below. 500 million or 5 billion tokens are used during continued pretraining.}
\label{tab:prelim_results}
\end{table}

\section{Watermarking with \method}



\begin{figure*}[t]
    \centering
    \includegraphics[width=0.9\linewidth]{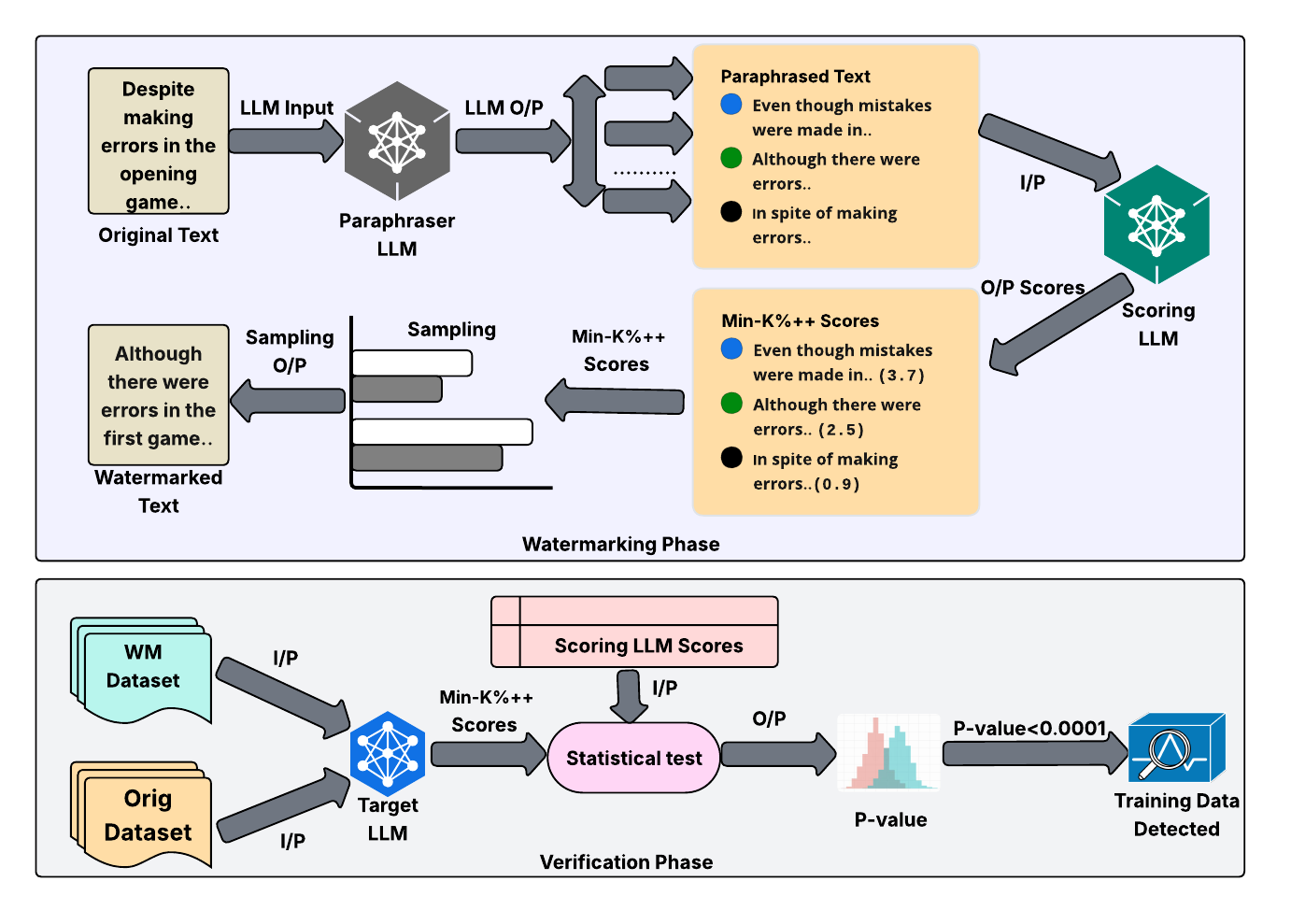}
    \caption{Overview of \method{}. \textbf{Watermarking Phase}: We use an LLM to generate multiple paraphrases of the original text. We sample one paraphrase that has a \minkpp score close to the original text. \textbf{Verification Phase}: Given a target LLM suspected of being trained on the watermarked data, we compute the \minkpp scores of the watermarked and original data and compare against the scores previously generated by the scoring model. Membership is detected through a paired t-test.}
    \label{fig:method}
\end{figure*}

This phase takes place after a content creator writes their content and before they publish it. To watermark a dataset, we generate multiple paraphrases of each document using a large language model and compute their \minkpp{} scores using a separate scoring model that has not been trained on the dataset being watermarked. This scoring model approximates the pre-training state of the target model. In practice, such a model is easy to find as \(D\) and \(D'\) are unpublished at the time of scoring. One can pick an open-source model with a knowledge cut-off prior to the release date of the dataset. We sample one paraphrase as the watermarked sample according to Algorithm 1.

The sampling favors paraphrases with scores close to the original score. For each side (above or below the original), we define a categorical distribution over candidate indices using weights proportional to \(exp(-\alpha|r_{ij}-1|)\) where \(\alpha>0\) controls the sharpness of the distribution. A higher 
\(\alpha\) causes the algorithm to favor paraphrases with scores close to the original more strongly. In practice, we pick the largest \(\alpha\) that does not lead to numerical underflow issues (\(\alpha = 100\)).

If both above- and below-original paraphrases are available, the side is chosen probabilistically in inverse proportion to how often each type appears globally across the dataset, helping to avoid systematic score shifts.
This sampling strategy ensures that the distribution of scores for the watermarked dataset remains similar to the original, reducing the likelihood of false positives when testing against models not trained on \(D'\). Importantly, because the \minkpp{} scores of watermarked text tend to increase after training, the score distribution after training becomes distinguishable from the pre-training state. Notably, if paraphrases were consistently selected only from the high-score side, it would create a detectable signature even without training—leading to false positives. \method{} avoids this by balancing selection, ensuring a reliable signal only when training has occurred.

\begin{algorithm}[!t]
\caption{Sampling paraphrases}
\begin{flushleft}
\textbf{Input:} Original scores \(\{s_i^{(0)}\}_{i=1}^{N}\), paraphrased scores \(\{S_i = \{s_{i1}, \dots, s_{im}\}\}_{i=1}^{N}\), paraphrases \(\{T_i = \{t_{i1}, \dots, t_{im}\}\}_{i=1}^{N}\), parameter \(\alpha=100\)\\
\textbf{Output:} Sampled paraphrases \(\{t_{ij_i}\}_{i=1}^{N}\)
\end{flushleft}

\begin{enumerate}
\item Pre-computation: define \( r_{ij} = s_{ij}/s_i^{(0)} \), then  
\(\mathcal{A} \!=\! \{ i: r_{ij}<1, \forall j\},\ \mathcal{B} \!=\! \{ i: r_{ij}>1, \forall j\}\).

\item Global side-balance:  \\
\(\pi_+ = \begin{cases}
0.5 & \text{if } |\mathcal{A}|+|\mathcal{B}|=0\\
\frac{|\mathcal{A}|}{|\mathcal{A}|+|\mathcal{B}|} & \text{otherwise}
\end{cases},
\\
\quad \pi_- = 1-\pi_+.\)

\item For each datapoint \(i=1,\dots,N\):
\begin{enumerate}
  \item Partition: \(\mathcal{R}_i^{(-)}=\{j: r_{ij}<1\},\ \mathcal{R}_i^{(+)}=\{j: r_{ij}>1\}\).
  \item Side \(s_i\): if one set empty, choose the other; else sample from \(\{+,-\}\) w.p. \(\pi_+,\pi_-\).
  \item Let \(\mathcal{R}=\mathcal{R}_i^{(s_i)}\).
  \item Weights: \(w_{ij}=\exp(-\alpha|r_{ij}-1|)\) for \(j\in\mathcal{R}\).
  \item Normalize: \(w_{ij}\leftarrow w_{ij}/\sum_{k\in\mathcal{R}} w_{ik}\).
  \item Sample \(j_i\) from categorical\(\{w_{ij}\}_{j\in\mathcal{R}}\).
\end{enumerate}
\item Return \(\{t_{ij_i}\}_{i=1}^N\).
\end{enumerate}
\end{algorithm}


\noindent

\section{Verification with \method{}}

During this phase, the content is released to the public and it is suspected that the data may have been used in an unauthorized manner for training.
Given an original document \( x \in D \) and its watermarked counterpart \( x' \in D' \), we define the score ratio under a model \(M\) as 
\[
r(x, x'; M) = \frac{f_{\text{Min-K\%++}}(x'; M)}{f_{\text{Min-K\%++}}(x; M)}.
\]
Let \(M_S\) denote the scoring model (not trained on \(D'\)),  \(M_T\) denote the target model (potentially trained on \(D'\)), and \(M_U\) denote the checkpoint of \(M_T\) before it was trained on \(D'\). Given that \minkpp scores rise after training, we can write \(\mathbb{E}_{x' \in D'}[f_{\text{Min-K\%++}}(x'; M_{T})]>\mathbb{E}_{x' \in D'}[f_{\text{Min-K\%++}}(x'; M_{U})]\). However, in practice, we do not have access to \(M_{U}\) and so we approximate it using \(M_{S}\). However, because \(M_{S}\) and \(M_T\) may differ in architecture or baseline predictions, we normalize each watermarked score by the corresponding original document score to allow meaningful comparisons between the two models. Thus, under the null hypothesis \(H_0\), the ratio of scores under \(M_T\) is equal to that under \(M_S\) when \(D'\) is not used for training \(M_{T}\), i.e., 
\[
H_0: \mathbb{E}_{x, x'}[r(x, x'; M_T)] = \mathbb{E}_{x, x'}[r(x, x'; M_S)].
\]
where the expectation is taken over \(x \in D\) and \(x' \in D'\). The alternate hypothesis \(H_1\) states that the ratio of scores under the target model is lower relative to the scoring model when \(D'\) was used for training, i.e., 
\[
H_1: \mathbb{E}_{x, x'}[r(x, x'; M_T)] < \mathbb{E}_{x, x'}[r(x, x'; M_S)].
\]
where the expectation is taken over \(x \in D\) and \(x' \in D'\). Note that the inequality flips sign here as \minkpp{} values in the denominator are always negative.
We test these hypotheses by computing these ratios for all pairs \((x, x')\) in each dataset and performing a 1-sided paired t-test.
A low p-value would indicate rejection of the null hypothesis \(H_0\), providing statistically significant evidence that the target model \(M_T\) has indeed been trained on the watermarked dataset \(D'\).

\section{Results}

\subsection{Datasets}
We ensure that the datasets we use for training have not previously been used to train our target model of interest. Thus we are limited to training models where the training data is known transparently. Two prominent model families that meet this criterion are the Pythia models and the OLMo \cite{groeneveld-etal-2024-olmo} models along with their corresponding training datasets, The Pile and Dolma.

\begin{enumerate}

\item \textbf{The Pile}: We use the deduplicated subsets of the Pile \cite{gao2020pile, duan2024membership} from the domains of Wikipedia (Wiki), Hackernews (HN), and Pubmed Central abstracts (PubMed) that were held out from training. 

\item \textbf{Dolma}: We use the PeS2o held-out subset of Dolma obtained from Paloma \cite{soldaini2024dolma, magnusson2024paloma}. All text in this subset was released after the release date of the Pile, making it non-member for Pythia models.

\end{enumerate}

\subsection{Models}
The watermarking pipeline and evaluation consists of 3 different types of models:

\begin{enumerate}

\item \textbf{Paraphraser model}: We use the Llama 3.1-405b model and generate 10 paraphrases per document \cite{grattafiori2024llama}.

\item \textbf{Scoring model}: We use the Pythia 2.8b-deduped model for the Pile datasets. We use a model that is known not to have been trained on our datasets, as otherwise, the distribution of the \minkpp{} scores would shift higher. Pythia 2.8b-deduped has different weights but is from the same model family as our target model, i.e., Pythia 410m. Consequently, for PeS2o we use the OLMo-1b model to investigate the effect of using a different model architecture between the scoring and target model.

\item \textbf{Target model}: This is the model that we suspect has been trained using \(D'\). We use the vanilla Pythia 410m model and Pythia 410m on which we do continued pretraining using watermarked datasets, as target models. During continued pretraining, in addition to the watermarked text, we sample 5 billion tokens of text from the Common Pile dataset \cite{kandpal2025common}.

\end{enumerate}




\subsection{Baselines}

We adopt the following baselines.
\begin{enumerate}
    \item \textbf{Maximum}: pick the paraphrase with the highest \minkpp score.
    \item \textbf{Random}: pick one of the paraphrases randomly.
\end{enumerate}

Additionally, \textbf{STAMP} and \textbf{LLM-DI} are described in the appendix.

\subsection{Evaluation of Watermarking}

\begin{table*}[htbp]
\centering
\begin{tabular}{l|ccc|ccc|ccc|ccc}
\toprule
\multirow{2}{*}{\textbf{Method}} & \multicolumn{3}{c|}{\textbf{PubMed}} & \multicolumn{3}{c|}{\textbf{Wiki}} & \multicolumn{3}{c|}{\textbf{HN}} & \multicolumn{3}{c}{\textbf{PeS2o}} \\
\cmidrule{2-13}
 & M & NM & NM/M & M & NM & NM/M & M & NM & NM/M & M & NM & NM/M \\
\midrule
LLM-DI   & 0.06 & \textbf{0.48} & 7.67 & 0.02 & \textbf{0.44} & 22 & 0.49 & \textbf{0.35} & 0.71 & 0.02 & \textbf{0.17} & 8.50 \\
STAMP    & 0.01 & \textbf{0.48} & 48 & 0.17 & \textbf{0.03} & 0.19 & 7E-4 & \textbf{0.15} & 214 & 0.15 & \textbf{0.46} & 3.07 \\
Maximum  & 0.03 & \textbf{1.00} & 3.33 & 1.00 & \textbf{1.00} & 1 & \textbf{3E-6} & \textbf{1.00} & 3E5 & 0.95 & \textbf{1.00} & 1.05 \\
Random   & \textbf{1E-7} & \textbf{8E-4} & 8E3 & \textbf{5E-9} & 2E-5 & 4E3 & \textbf{4E-27} & \textbf{0.10} & 3E25 & 1E-3 & \textbf{0.11} & 100 \\
SPECTRA  & \textbf{1E-17} & \textbf{0.02} & 2E15 & \textbf{4E-19} & \textbf{0.02} & 5E16 & \textbf{3E-60} & \textbf{0.59} & 2E59 & \textbf{2E-12} & \textbf{3E-3} & 2E9 \\
\bottomrule
\end{tabular}
\caption{p-values for different baselines compared against \method. Bold indicates a statistically significant result for detecting membership under a threshold of \(p<10^{-4}\) for members (M) using Pythia 410m trained on the watermarked data as the target model. Note that the paraphrases for datasets used during continued pretraining (meant for detection) are different for each row and thus result in different target models. Bold also indicates that for non-members (NM), the p-value was above the threshold, indicating that membership was not falsely detected using Pythia 410m as the target model.}
\label{tab:main_results}
\end{table*}
We measure statistical significance (p-values) for detecting the watermark in each dataset. Specifically, we compute member p-values from the Pythia-410m model trained on the watermarked datasets and non-member p-values from the original Pythia 410m model that has not encountered the watermarked data during training.

We see from the results (Table \ref{tab:main_results}) that \method{} is the \textit{only one} that correctly detects membership for each dataset in the study under a threshold of \(p<10^{-4}\). Under a naïve approach of selecting paraphrases that maximize the \minkpp shift, the resulting pre-training shift is so large that subsequent training does not further amplify it, making pre- and post-training distributions indistinguishable and thus undetectable in practice. The random baseline also fails as non-member Wiki results in false positives, while with PeS2o, it fails to detect membership. This indicates that the sampling strategy employed for paraphrases in \method{} is crucial to ensuring its performance. \method{} consistently has a high ratio of p-value between member data and non-member data (\(>10^{9}\)) which is higher than the baselines. Notably, \method{} correctly detects membership for PeS2o, demonstrating that \method{} remains effective even as the scoring model and target model architecture differ. As suggested in \citet{huh2024position}, different LLMs follow similar training objectives, and as the amount of training data and tasks gets scaled up, the space of acceptable representations narrows dramatically, leading to similar learned representations.
 
In contrast, STAMP and LLM-DI fail to reliably detect members under our strict threshold (\(p<10^{-4}\)). Although these methods achieve significance for certain datasets when adopting a more permissive threshold \(p<0.05\), we argue that due to the significant implications of falsely identifying datasets as training data in LLMs, a more stringent threshold is necessary and justified.

\subsection{Validating Quality of Paraphrases}
We validate the quality of paraphrasing by using the P-SP metric \cite{wieting-etal-2022-paraphrastic}. The P-SP metric is widely used to measure paraphrasing quality \cite{rastogi2025stampcontentprovingdataset, krishna2023paraphrasing}. For a human-generated paraphrase, the average P-SP is 0.78 \cite{krishna2023paraphrasing}. Except for Hackernews, all watermarked datasets had a P-SP score above 0.88 (Table \ref{tab:p-sp_scores}). The lower scores for Hackernews are explored next.

\begin{table}
\centering
\begin{tabular}{@{}llcccc@{}}
\toprule
& \textbf{PubMed} & \textbf{Wiki} & \textbf{HN} & \textbf{PeS2o} \\
\midrule
P-SP & 0.88 & 0.93 & 0.76 & 0.93 \\
\bottomrule
\end{tabular}
\caption{P-SP scores on paraphrasing quality. P-SP scores measure how well the paraphrase preserves the semantic content of the original document.}
\label{tab:p-sp_scores}
\end{table}

\noindent\textbf{Human evaluation}. 
We randomly select 54 watermarked documents from our datasets and distribute them among four evaluators. This distribution ensures that each evaluator reviews 27 documents, with each document being assessed by two different evaluators. The evaluation focuses on whether the paraphrasing preserves the (i) meaning, (ii) structure, and (iii) author tone of the original text. Evaluators evaluate on a Likert scale \cite{likert1932technique} of 1-5, with 5 being the best and 1 the worst.

\begin{figure}
    \centering
    \includegraphics[width=0.9\linewidth]{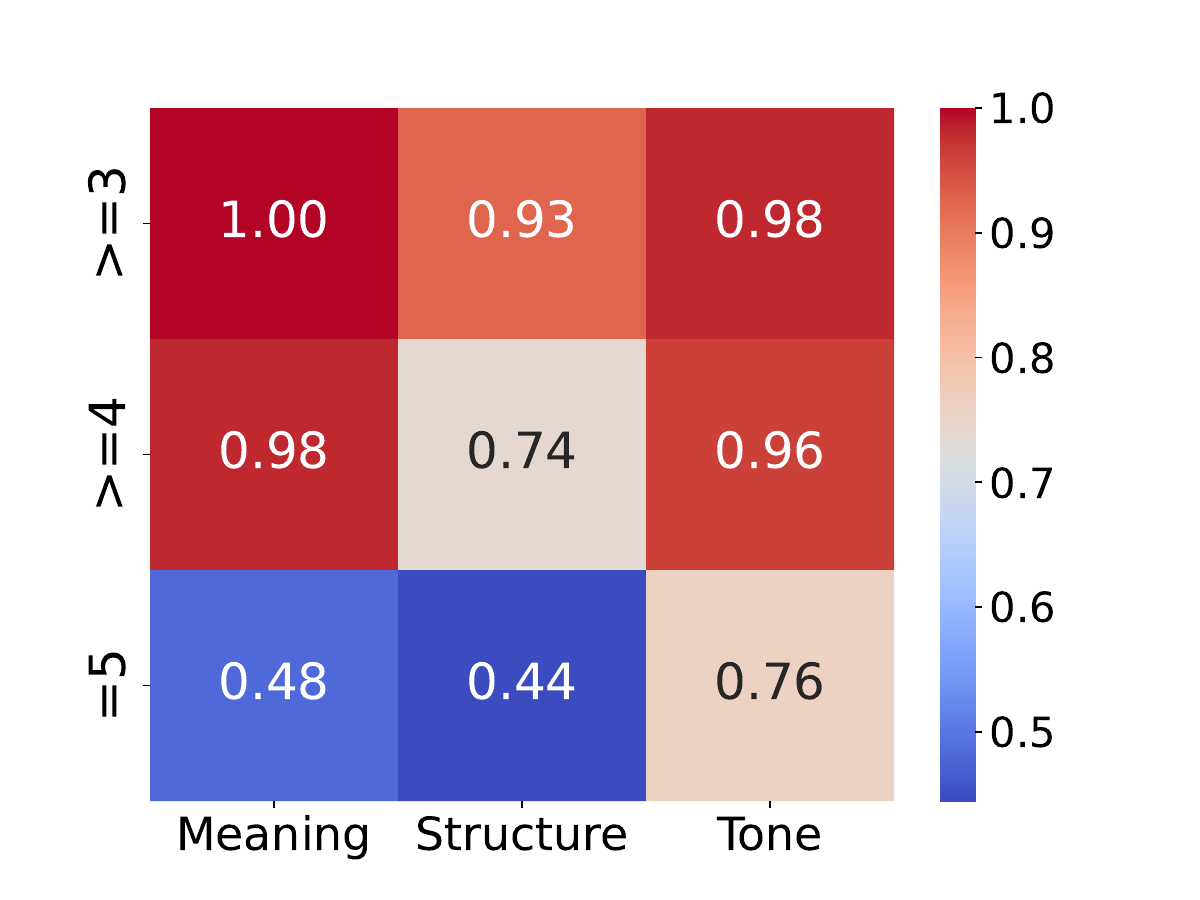}
    \caption{Heatmap showing fraction of evaluator scores for which the paraphrases received a rating \(\geq 3\).}
    \label{fig:human_eval}
\end{figure}

Figure \ref{fig:human_eval} shows the fraction of points that achieved a mean evaluator score of \(\geq 3\). While the mean scores for all three criteria exceeded 4, the scores for the structure preservation criterion are comparatively lower. Specifically, for conversational-style text such as on Hackernews, the paraphraser LLM occasionally fails to maintain the original structure.

\section{Ablation Studies}

\subsection{Number of Samples}

\begin{figure}[!ht]
    \centering
    \includegraphics[width=1.0\linewidth]{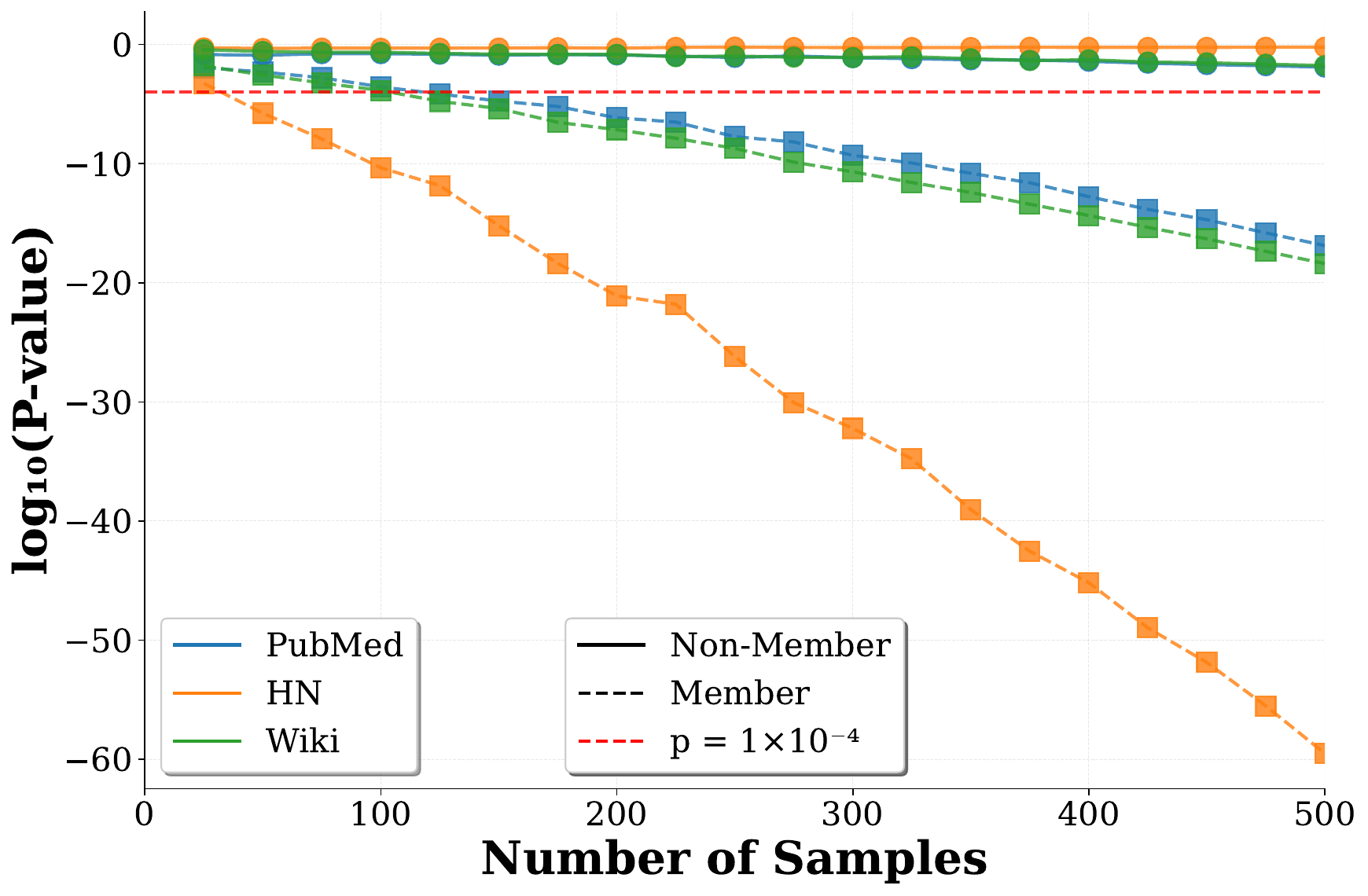}
    \caption{Trend of p-values with number of samples}
    \label{fig:p_value_samples}
\end{figure}

The p-value for any of the datasets we test goes below the threshold after 100-150 samples, suggesting that this is the minimum number of samples needed (Figure \ref{fig:p_value_samples}). The p-value for non-members is always above the threshold.

\subsection{Ablations on Scoring Model}
We investigate the effect of using different scoring models for the ranking of paraphrases. We aim to ascertain whether changing the scoring model would substantially affect the outcomes of \method{}. We utilize paraphrases derived from the PeS2o dataset. The OLMo-1b model was originally employed for scoring the PeS2o dataset in our experiments. We compare the rank ordering of paraphrases generated by OLMo-1b against OLMo-7b, Pythia 2.8b, Pythia 160m, and Pythia 6.9b. Each of these models, along with the original scoring model, is used to compute \minkpp{} scores for the paraphrases from the PeS2o dataset. The rankings generated by each model are then used to compute the Spearman~\cite{spearman2010proof} and Kendall~\cite{kendall1938new} rank correlation scores between the original scoring model (OLMo-1b) and the additional scoring models.
\begin{table}[ht]
    \centering
    \begin{tabular}{lcc}
        \toprule
        Additional Model & Spearman $\rho$ & Kendall $\tau$ \\
        \midrule
        Olmo-7b    & 0.826 & 0.639 \\
        Pythia-2.8b-deduped  & 0.824 & 0.635 \\
        Pythia-160m-deduped & 0.699 & 0.514 \\
        Pythia-6.9b  & 0.818 & 0.631 \\
        \bottomrule
    \end{tabular}
    \caption{Rank-correlation coefficients between OLMo-1b and other scoring models.}
    \label{tab:rank_corr}
\end{table}

For three out of the four additional models, the Spearman's rank correlation coefficient ($\rho$) values exceed 0.8 (Table \ref{tab:rank_corr}). The correlation for the Pythia 160m model is lower than that of others. The correlation is above 0.8 for models with more than 2.8b parameters, indicating that the correlation stabilizes for larger models. For Kendall's $\tau$, a value of \(>0.6\) is considered a strong agreement~\cite{lyu2021towards}. For three out of four datasets, the  $\tau$ score \(>0.6\).
These findings suggest that \method{} is robust to changing the scoring model as long as the model is sufficiently capable.




\section{Conclusions}

We presented \method, a watermarking approach that enables content creators to test if their data was used to train an LLM. Unlike previous approaches, \method{} does not need access to the decoding layer of a large LLM. \method{} also does not require access to a held-out dataset from the same domain that is typically necessary for MIA. We empirically show that \method{} can achieve a p-value gap of at least nine orders of magnitude between member and non-member data with 500 samples, making this a reliable test of membership.

We highlight some important limitations of our study. We continue to pre-train the Pythia 410m model over a large number of tokens instead of training an LLM from scratch due to computational constraints. We assume access to the log probabilities output by the model, which can be challenging to achieve for proprietary models. Our approach will not help content creators who have already published their content, but can only help them going forward, as the watermarking must be done before publishing. For text data with some structure in it, such as conversational text, our paraphrasing approach does not perform very well.

In terms of future directions, there is a need to develop watermarking techniques that can be used to verify membership not only by the content creator but also by interested third parties. Additional verification of our techniques at the scale of pre-training a model would inspire greater confidence that it can be employed in the event of any legal challenges.

\section*{Acknowledgements}
This paper was prepared for informational purposes by the Artificial Intelligence Research group of JPMorgan Chase \& Co. and its affiliates (``JP Morgan'') and is not a product of the Research Department of JP Morgan. JP Morgan makes no representation and warranty whatsoever and disclaims all liability, for the completeness, accuracy, or reliability of the information contained herein. This document is not intended as investment research or investment advice, or a recommendation, offer or solicitation for the purchase or sale of any security, financial instrument, financial product, or service, or to be used in any way for evaluating the merits of participating in any transaction, and shall not constitute a solicitation under any jurisdiction or to any person, if such solicitation under such jurisdiction or to such person would be unlawful.

\bibliography{aaai2026}

\section{Appendix}

\subsection{Datasets: Additional details}

For each domain in the Pile that we use, we sample 500 documents for watermarking, and 500 are kept as held-out non-member data that are not used for training and are instead used as the validation set for LLM-DI. We pick the datasets from \citet{duan2024membership} that were deduplicated against the Pile training data with a 13-gram Bloom filter and a threshold of 80 \% overlap. This means that the dataset was constructed from the held-out set of the Pile to avoid any document with up to a 13-gram overlap with any document included in the training set. Each training document was limited to between 100 and 200 words and at most 512 tokens. We control for the length as longer sequences make the document more detectable to MIA methods \cite{puerto-etal-2025-scaling}.\\

\noindent\textbf{PeS2o}: We use the PeS2o subset of Dolma available in the Paloma dataset \cite{magnusson2024paloma}. To further deduplicate it, we compare the `ID' field of this dataset against the entire Dolma dataset and remove all documents from Paloma whose `ID' was found in Dolma. This left us with 432 documents, which we truncated as above to 512 tokens with between 100 and 200 words. We further split this into 216 documents used for watermarking and 216 documents used as held-out data.

The number of tokens in each dataset is shown in Table \ref{tab:num_tokens}

\begin{table}[htbp]
\centering
\begin{tabular}{@{}lc@{}}
\toprule
\textbf{Dataset} & \textbf{Number of tokens} \\
\midrule
Pubmed Central & 154387 \\
Wikipedia & 141450 \\
Hackernews & 160637 \\
PeS2o & 56770 \\
\bottomrule
\end{tabular}
\caption{Number of tokens for each dataset computed using the Pythia tokenizer}
\label{tab:num_tokens}
\end{table}

\subsection{Paraphraser LLM}

We sample 10 unique paraphrases using the Llama 3.1-405b model using the prompt provided below. Each paraphrase is sampled at a different temperature value. We verify that each paraphrase is unique after lower-casing it. For cases where the generated paraphrase matches a previously generated paraphrase for the same document, we repeat the generation until we obtain a unique paraphrase.


\tcbset{
  mypromptbox/.style={
    enhanced,
    colback=gray!5,
    colframe=gray!50!black,
    boxrule=0.8pt,
    arc=4pt,
    outer arc=4pt,
    boxsep=5pt,
    left=5pt,
    right=5pt,
    top=5pt,
    bottom=5pt,
    fonttitle=\bfseries,
    title=SPECTRA Paraphrasing Prompt,
    coltitle=white,
    sharp corners=south,
    breakable,
    listing only,
    listing options={
      basicstyle=\ttfamily\small,
      breaklines=true,
      columns=fullflexible
    }
  }
}


\begin{tcolorbox}[mypromptbox]
Paraphrase the below paragraph of text enclosed in \textless text\textgreater tags, hereafter referred to as the original text. \
The paraphrased text should use different vocabulary, sentence structure, and style while preserving the meaning and tone of the original text. \
Do not remove any information from the original text while paraphrasing. \
Do not add any new information to the paraphrased text that is not present in the original text. \
Do not add any interpretive language to the paraphrased text that is not implied by the original text. \
Ensure that all technical details, findings, results, and other information such as tense, voice, and line breaks are preserved. \
Format your response as: PARAPHRASED PARAGRAPH: [your rephrased version] \
Based on the aforementioned directions, paraphrase the following text. \textless text \textgreater \{text\}\textless /text \textgreater 
\end{tcolorbox}

\subsection{Continued Pretraining}

In addition to the watermarked datasets, we sample documents from the Common Pile dataset for continued pre-training of Pythia 410m. Specifically, we select documents from the USPTO, USGPO, ArXiv, LibreText, and Doab domains of the Common Pile \cite{kandpal2025common}. Each document is split into sequences of 512 tokens and used for training after randomizing the order. To avoid overlap with data seen during initial pre-training, we only sample documents published after December 2020, which is the cut-off date of the Pile dataset.

We use the AdamW optimizer with a learning rate of \(10^{-4}\), \((\beta_1, \beta_2) = (0.99, 0.999)\), cosine decay and a batch size of 40. A warmup of 0.5\% of training tokens was used with no weight decay. We trained all our models on an L40S Tensor Core GPU. We used the Transformers library (v4.43). We use a random seed of 1234 for all our experiments.

\subsection{Additional Details of MIA methods}

\textbf{DC-PDD}: This method quantifies the divergence between the token log probabilities predicted by a target model and the empirical probabilities of those tokens in a reference corpus. Formally, we compute
\(
P_M(x_t \mid x_{<t})
\)
where $P_M(x_t \mid x_{<t})$ is the model's predicted probability distribution for token $x_t$ and
\(
Q_{\text{ref}}(x_t)
\)
where $Q_{\text{ref}}(x_t)$ is the marginal token-frequency distribution estimated from the reference corpus, independent of context. $Q_{\text{ref}}(x_t)$ is pre-computed by adding the counts of each token in the vocabulary and computing its frequency against the total number of tokens in the corpus.

For each document \(x=(x_1,\dots,x_n)\), DC‑PDD computes a calibrated score by measuring the divergence between the model’s token‐probability distribution and the reference token‐frequency distribution. Specifically, it evaluates the following cross‐entropy score:
\[
 -\frac{1}{|FOS(x)|} \sum_{t\,:\,\text{first occurrence}}  P_M(x_t \mid x_{<t}) \cdot \log Q_{\text{ref}}(x_t)
\]
where \(FOS(x)\) denotes the set of tokens in $x$ at their first occurrence. Only the first occurrence of each token in \(x\) is considered to reduce bias from repeated exposures.

For Min-K\% and \minkpp, we used K=20 \%, which is the standard setting used in \citet{duan2024membership}. We used the Mimir library (https://github.com/iamgroot42/mimir) to compute all MIA scores.

\subsection{Watermarking baselines}

\begin{enumerate}
    \item \textbf{STAMP} \cite{rastogi2025stampcontentprovingdataset}: Text is watermarked using the KGW scheme by rephrasing it with a Llama-70B. This rephrasing is the public version. The same text is also watermarked using several other keys that are kept private. The perplexity is computed using the target model for all rephrases. The perplexity of the public watermarked text sequence is compared against the average perplexity of the private watermarked sequences over the dataset using a paired t-test. If the watermarked text is used for training the target model, then it is expected to have lower perplexity than the average of the private rephrases. The implementation provided by the authors was used in our study.

    In the evaluation of STAMP, the authors mainly considered Question Answering (QA) datasets; hence, in the prompt, the wording `rephrase the question' is used. As we do not use QA datasets, we have modified the prompt to `rephrase the text'.
    
    \item \textbf{LLM-DI}: LLM-DI \cite{mainidi2024} utilizes outputs from multiple MIA methods as features to train a classifier for detecting training data. In this setting, the content creator retains an additional unreleased dataset. To ascertain whether a model has been trained on a released dataset, the creator must generate features from both the released and unreleased datasets and subsequently train a classifier using these features. In our experimental setup, each dataset was divided into two equal parts, with a 50:50 split to obtain training and validation datasets. The implementation provided by the authors of LLM-DI was utilized in our study. Note that the original datasets were used during continued pretraining and for detection without introducing any watermarks.
\end{enumerate}

\tcbset{
  mypromptbox/.style={
    enhanced,
    colback=gray!5,
    colframe=gray!50!black,
    boxrule=0.8pt,
    arc=4pt,
    outer arc=4pt,
    boxsep=5pt,
    left=5pt,
    right=5pt,
    top=5pt,
    bottom=5pt,
    fonttitle=\bfseries,
    title=STAMP Paraphrasing Prompt,
    coltitle=white,
    sharp corners=south,
    breakable,
    listing only,
    listing options={
      basicstyle=\ttfamily\small,
      breaklines=true,
      columns=fullflexible
    }
  }
}

\begin{tcolorbox}[mypromptbox]
Rephrase the text given below. Ensure you keep all details present in the original, without omitting anything or adding any extra information not present in the original text.

Text: ```text'''

Your response should end with Rephrased Text: [rephrased text] 
\end{tcolorbox}

\subsection{Additional results}


\subsubsection{Non-member analysis}
We computed the p-value of PeS2o on other LLMs as the target model that had a knowledge cut-off prior to the earliest released samples in PeS2o and thus could not have been trained on it (Table \ref{tab:false_positives}). All computed p-values are greater than our threshold of $10^{-4}$. Thus \method{} did not generate any false positives in this test.


\begin{table}[htbp]
\centering
\begin{tabular}{@{}lc@{}}
\toprule
\textbf{Model} & \textbf{p-value} \\
\midrule
Mistral-7b & 0.08 \\
GPT-Neo-2.7b & 0.04 \\
GPT-J-6b & 0.003 \\
Bloom-3b & 0.33 \\
Bloom-7b & 0.02 \\
\bottomrule
\end{tabular}
\caption{p-value of PeS2o on models that have not seen it during pre-training. All models have \(p>10^{-4}\), indicating that membership could not be detected with statistical significance. OLMo-1b is the scoring model.}
\label{tab:false_positives}
\end{table}

\subsubsection{Watermarking Evaluation via Membership Inference Attacks}

In addition to using p-values for watermark evaluation in the main paper, we report here the results of evaluating watermarking through a conventional Membership Inference Attack (MIA) framework. In this setting, for each dataset, we compare scores computed for documents that were used to train the model (“member” data) against scores for documents that were not included in training (“non-member” data). These scores are then used as inputs to a binary classifier that predicts whether a document is a member or a non-member. Member and non-member data are expected to have separable score distributions.

Specifically, for the Pythia 410 model trained with data watermarked using \method{}, we calculate the area under the Receiver Operating Characteristic curve (ROC-AUC) and the true positive rate (TPR) at a fixed false positive rate (FPR) of 1\%. High ROC-AUC values indicate strong separation between member and non-member data, while high TPR at low FPR measures the ability to detect true positives with minimal false positives.

As shown in Table~\ref{tab:roc_auc_after_training}, \minkpp achieves the highest ROC-AUC on three out of four datasets. However, as indicated in Table~\ref{tab:tpr_at_low_fpr_after_training}, even when ROC-AUC is high, reliably detecting members at a very low false positive rate remains challenging, highlighting a practical limitation of these approaches in real-world detection scenarios.



\begin{table}[htbp]
\centering
\begin{tabular}{l|cccc}
\hline
 & \textbf{PubMed} & \textbf{HN} & \textbf{Wiki} & \textbf{ PeS2o} \\
\hline
Min-K & 0.653 & 0.810 & 0.691 & 0.684 \\
Loss & 0.564 & 0.728 & 0.669 & 0.636 \\
DC-PDD & 0.663 & \textbf{0.856} & 0.709 & 0.710 \\
\minkpp & \textbf{0.743} & 0.855 & \textbf{0.761} & \textbf{0.718} \\
\hline
\end{tabular}
\caption{ROC AUC of member data against non-member data computed using Pythia 410m model trained on watermarked data}
\label{tab:roc_auc_after_training}
\end{table}

\begin{table}[htbp]
\centering
\begin{tabular}{l|cccc}
\hline
\ & \textbf{PubMed} & \textbf{HN} & \textbf{Wiki} & \textbf{PeS2o} \\
\hline
Min-K & 4.4 & 16.4 & 3.4 & 2.3 \\
Loss & 0.8 & 3.8 & 1.6 & 1.9 \\
DC-PDD & 4.8 & \textbf{27.2} & 4.0 & 5.1 \\
\minkpp & \textbf{7.2} & 24.0 & \textbf{4.2} & \textbf{9.3} \\
\hline
\end{tabular}
\caption{True Positive Rate (\%) at False Positive Rate 1 \% computed using Pythia 410m model trained on watermarked data}
\label{tab:tpr_at_low_fpr_after_training}
\end{table}

\subsubsection{Effect of Scoring Model on Statistical Testing}

In our primary experiments, we used the OLMo-1b model as the scoring model for watermarking the PeS2o dataset, and computed p-values via statistical testing on a target model. In Table 4, we investigated whether changing the scoring model during watermarking affected the ranking of paraphrases, finding that the order remained largely consistent across models. Here, we extend this analysis to examine the effect of varying the scoring model used during the statistical test itself. 
None of the scoring models employed for evaluation were exposed to the PeS2o dataset during their respective pre-training. As shown in Table~\ref{tab:scoring_model_p_value}, in all cases, PeS2o is detected with statistically significant p-values when it was included in the training data for the target model. Membership was not detected with statistical significance (\(p<10^{-4}\)) when using the vanilla Pythia 410m model (not trained on watermarked data) using any of the scoring models. This demonstrates the robustness of our approach to varying the scoring model in the statistical test.

\begin{table}[htbp]
\centering
\begin{tabular}{@{}lcc@{}}
\toprule
\textbf{Scoring model} & \textbf{Member} & \textbf{Non-member}\\
\midrule
OLMo-7b & 3E-5 &  0.56\\
Pythia-2.8b-deduped & 2E-7 & 0.08\\
Pythia-6.9b-deduped & 1E-5 & 0.59\\
OLMo-1b & 2E-12 & 3E-3\\
\bottomrule
\end{tabular}
\caption{Effect of varying the scoring model for PeS2o.}
\label{tab:scoring_model_p_value}
\end{table}

\subsection{Additional Ablation studies}

\subsubsection{Effect of training tokens on p-value}

\begin{figure}
    \centering
    \includegraphics[width=0.9\linewidth]{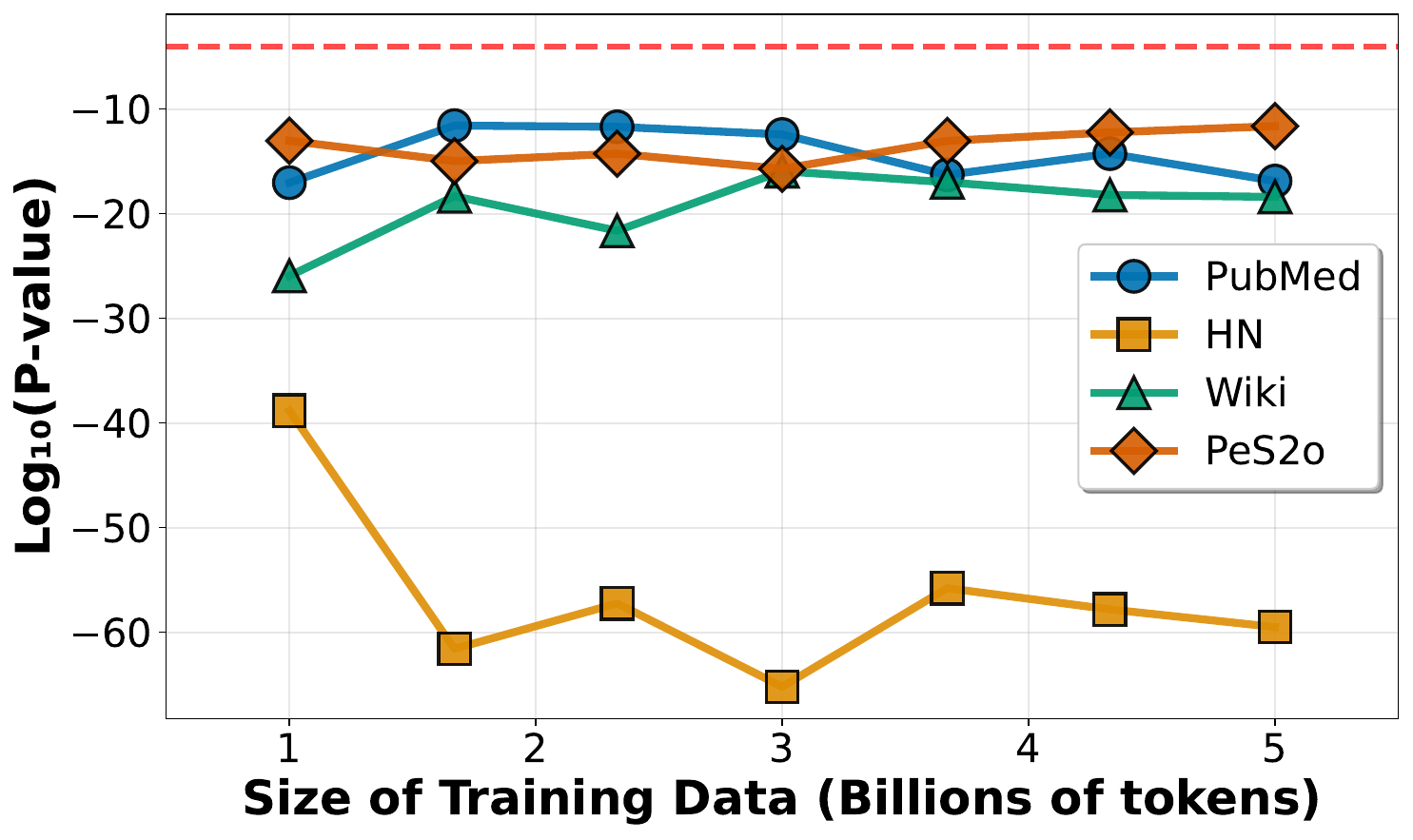}
    \caption{Trend of p-values with the number of pre-training tokens. The dashed red line indicates the p-value threshold of \(10^{-4}\)}
    \label{fig:pretraining_tokens}
\end{figure}

For all four datasets, the p-value remains well below the significance threshold \(p <10^{-4}\) throughout the training run (Figure~\ref{fig:pretraining_tokens}). This indicates that our t-test would be effective at any point during training. After 3.5 billion training tokens, the p-values show little further change, suggesting that the test results remain stable even as training progresses beyond this point.

\subsubsection{Evaluating Alternative MIA Scores in \method}

We replace the \minkpp{} score in \method{} with several other common MIA scores and report the resulting p-value performance in Table~\ref{tab:other_scores}. Most alternative methods either tend to produce false positives (such as Min-K\% and loss) or fail to reliably detect membership (such as DC-PDD).



\begin{table*}[htbp]
\centering
\begin{tabular}{l|cc|cc|cc|cc}
\toprule
\multirow{2}{*}{\textbf{Method}} & \multicolumn{2}{c|}{\textbf{PubMed}} & \multicolumn{2}{c|}{\textbf{Wiki}} & \multicolumn{2}{c|}{\textbf{HN}} & \multicolumn{2}{c}{\textbf{PeS2o}} \\
\cmidrule{2-9}
 & M & NM &  M & NM &  M & NM  & M & NM \\
\midrule
Loss   & 1E-73 & 4E-13 &  4E-34 & 6.2E-20 & 6E-178 & 1E-53 & 7E-23 & 7E-11 \\
DC-PDD    & 1.0 & 1.0 &  1.0 & 0.98 & 1.0 & 0.49 & 1.0 & 1.0 \\
Min-K\%  & 2E-74 & 2.8E-10 & 1.6E-34 & 1E-17 & 1E-169 & 3.3E-12 & 2E-21 & 4E-7 \\
\minkpp  & 1E-17 & 0.02 & 4E-19 & 0.02 & 3E-60 & 0.59 & 2E-12 & 3E-3 \\
\bottomrule
\end{tabular}
\caption{p-values for different scoring methods compared against \method{}. \(p<10^{-4}\) for membership to be detected with statistical significance. For all datasets, methods other than \minkpp are unable to either detect or reject membership, showing a consistent bias for one or the other depending on the method. M refers to member data, and NM refers to non-member data (target model Pythia 410m). The last row corresponds to \method{}.}
\label{tab:other_scores}
\end{table*}

\subsection{Human evaluation}
\tcbset{
  mypromptbox/.style={
    enhanced,
    colback=gray!5,
    colframe=gray!50!black,
    boxrule=0.8pt,
    arc=4pt,
    outer arc=4pt,
    boxsep=5pt,
    left=5pt,
    right=5pt,
    top=5pt,
    bottom=5pt,
    fonttitle=\bfseries,
    title=Instructions for Human Evaluation,
    coltitle=white,
    sharp corners=south,
    breakable,
    listing only,
    listing options={
      basicstyle=\ttfamily\small,
      breaklines=true,
      columns=fullflexible
    }
  }
}

\begin{tcolorbox}[mypromptbox]
Please rate the paraphrased text based on the following three criteria.

1. Meaning Preservation, i.e., Rate higher if the meaning of the original text is better preserved. Please do not consider emotional tone in this rating.

2. Structural Preservation, i.e., Rate higher if the structure of the original text is better preserved.

3. Emotional Tone Preservation, i.e., Rate higher if the tone of the original text is better preserved.

Please score on a scale of 1-5, with 1 being the worst and 5 being the best. 
 
\end{tcolorbox}
\noindent\textbf{Limitations of Paraphrasing}. During human evaluation, it was observed that paraphrasing sometimes altered the structure of the original content, especially for conversational text.
For example, in a sample from the Hackernews dataset (Figure~\ref{fig:struct1}), the LLM paraphrased a multi-speaker conversation about using Google Alerts to detect potential website hacking. However, the resulting paraphrased text did not preserve the conversational format, instead providing a summarized account of the discussion. This loss of dialogue structure may reduce the fidelity of paraphrased data for tasks that rely on conversational cues.


\begin{figure*}
\begin{tcolorbox}[title=Original and Paraphrased Text example for Hackernews dataset, enhanced, raster equal height, boxrule=0.6pt, colback=white]
{
  \textbf{Original}\par
  \smallskip
  \begin{quote}
  How to use Google Alerts to find out if your site gets hacked - greg
http://www.blogstorm.co.uk/how-to-use-google-alerts-to-find-out-if-your-site-gets-hacked/

======
pierrefar
It's a good backup solution as by the time Googlebot finds the cracked pages,
it's a bit too late.

I'd set up a cron job to check the database once a day or 12 hours or so.
That's more pre-emptive.

~~~
ZeroGravitas
I think you're missing the point. These SEO hackers don't announce themselves
by defacing your front page or deleting your database. They insert hidden
links in your HTML to boost their PageRank and it's in their interests to go
undetected for as long as possible.

~~~
pierrefar
I'm an SEO and know full well what they're doing. My daily cron job suggestion
is to detect their activities.

------
peregrine
You would need a slightly larger set of words, but nonetheless an interesting
solution.

~~~
streety
Absolutely. Perhaps harvest words from blog spam.
  \end{quote}
}
{
  \textbf{Paraphrased}\par
  \smallskip
  \begin{quote}
  To detect potential hacking of your website, consider utilizing Google Alerts. However, by the time Googlebot discovers compromised pages, it may be too late. A more proactive approach would be to set up a scheduled task to check the database regularly, such as daily or every 12 hours. This method can help identify malicious activities before they cause significant damage. It's important to note that SEO hackers often insert hidden links in HTML to boost their PageRank, aiming to remain undetected for as long as possible. To combat this, a daily database check can be an effective solution. Additionally, using a larger set of words, possibly harvested from blog spam, can enhance the detection process.

  \end{quote}
}
\end{tcolorbox}
\caption{Original and Paraphrased Texts from the Hackernews Dataset: Exploring Methods to Detect Website Hacking Using Google Alerts. In this example, paraphrased example does not follow the conversational format of the original text.}
\label{fig:struct1}
\end{figure*}

\subsection{P-SP metric}

The P-SP (PARAGRAM-SP) metric is used to quantify the semantic similarity between an original text and its paraphrase \cite{wieting-etal-2022-paraphrastic}. P-SP is computed as the cosine similarity between embeddings of the original and paraphrased texts, where each embedding is obtained by averaging the sentencepiece token embeddings of the respective text. In this metric, the only learned parameters are the sentencepiece embeddings themselves, which are trained on the ParaNMT paraphrase corpus \cite{wieting2018paranmt}.

\subsection{Extended Related Work}

 \paragraph{Membership Inference Attack} Membership inference attacks (MIA) were first systematically studied for simpler neural and convolutional networks by \citet{shokri2017membership}. Their approach involves training multiple shadow models, each on different non-overlapping subsets of data drawn from the same distribution as the target model's training set. The shadow models generate features used to train an attack model that predicts membership status. However, this shadow model approach is computationally infeasible for large language models (LLMs) with billions of parameters.

Subsequent work has focused on more scalable membership inference methods. For example, \citet{mattern-etal-2023-membership} proposed the Neighborhood Attack, which perturbs a document by masking parts of its text and replacing the masked tokens with predictions from a BERT-like model (``neighbor"). The MIA score is defined as the difference between the loss on the original document and the average loss across its perturbed neighbors; training documents typically yield lower values than non-members.

The MIA techniques described thus far do not require a reference model. In contrast, reference-based attacks such as LiRA (Likelihood Ratio Attack) \cite{mireshghallah2022quantifying} do require a reference model trained on data from a similar, but largely non-overlapping, distribution. LiRA measures membership by comparing the loss of the target model to that of the reference model on the same document, thereby calibrating for intrinsic document difficulty. As observed by \citet{duan2024membership}, obtaining a suitable reference model for LiRA can be challenging in practice.

Other recent works have shown that knowledge of non-member data can be leveraged to calibrate detection. \citet{bertran2023scalable} use quantile regression on the likelihoods of non-member data (computed with the target model) to estimate a threshold separating members from non-members. Unlike earlier work, which often focuses on detecting pre-training data (where the model sees each example only once), Bertran et al.\ target fine-tuning data, which may be seen multiple times during training.

More recently, the ReCALL method prepends a prefix of non-member data to member documents and computes the ratio of the document likelihood with and without the prefix. They demonstrate that this likelihood ratio changes more for member documents than for non-members \cite{xie-etal-2024-recall}.

While the term ``attack" is often used in an adversarial context, denoting unauthorized detection of training data, in our work, we repurpose MIA techniques as tools for data transparency—enabling enforcement of data use policies and verification of model training practices.

\paragraph{Training Data Privacy} A related line of work, known as training data extraction, seeks to prove membership by generating verbatim training examples from a trained model~\cite{carlini2021extracting}. However, successful extraction typically requires knowledge of effective prefixes or prompts and a reliable method for verifying whether the generated output indeed originated from the training set. Moreover, not all data used to train a model is equally susceptible to extraction, as memorization varies widely across training examples.

Differential privacy is a technique designed to limit the information leakage of individual training data points. This is usually achieved by clipping gradients on a per-example basis, thus bounding the contribution of any single example to the model parameters. However, differential privacy methods often result in reduced downstream performance and are rarely adopted in large language model (LLM) training, since per-example gradient computation significantly increases training costs~\cite{li-etal-2024-fine}.

Beyond membership inference attacks, alternative signals have been proposed to detect whether a particular dataset was used in model training. For example, \citet{oren2023proving} investigate whether a model assigns higher likelihood to the canonical ordering of documents in a dataset compared to other permutations. This may indicate exposure to the dataset during training. However, this method relies on the assumption that data are presented in canonical order during training, which may not hold in practice due to common data shuffling procedures.

 


 \paragraph{Watermarking LLM output} While our work focuses on watermarking training data, a large body of literature addresses watermarking the output of LLMs to detect whether an LLM generated a given text. A popular statistical watermarking technique partitions the model's vocabulary into a “red list” and a “green list,” and increases the logits of the green list tokens by a constant $\delta$ during generation. The presence of a watermark is then detected by measuring the frequency of green list tokens in the output and applying a statistical test, such as a z-test~\cite{kirchenbauer2023reliability}.

\citet{sander2024watermarking} demonstrated that watermarked text is ``radioactive'' meaning it leaves a detectable trace if the watermarked text is later used to train another model. \citet{sanderdetecting} extended this technique to detect benchmark contamination in LLMs, but found that successful detection required the contaminated data to be repeated many times in the pretraining corpus (at least 16 times). \citet{rastogi2025stampcontentprovingdataset} reported that the statistical test from \citet{sanderdetecting} was unable to detect training data when it appeared fewer times in the corpus. Additionally, \citet{panaitescu2025can} found that watermarking text can diminish the effectiveness of membership inference attack (MIA) techniques.

\section{Watermarking Examples}

We provide an example from each of our datasets of the watermarked and original document for \method{} in Figure 6-9.


\begin{figure*}
\begin{tcolorbox}[title=Original and Paraphrased Text example for Wikipedia dataset, enhanced, raster equal height, boxrule=0.6pt, colback=white]
{
  \textbf{Original}\par
  \smallskip
  \begin{quote}
  Charles Theodore 'Theo' Harding (26 May 1860 – 13 July 1919) was an English-born international rugby union player who played club rugby for Newport and international rugby for Wales. Harding was an all-round sportsman and also captained Newport Hockey Club in their very first season.

Rugby career

Harding was one of the first Newport players and was given the captaincy of the club in the 1887/88 and the 1888/89 season. During the 1888 Harding twice faced the first overseas touring team the New Zealand Māoris. The first occasion was also Harding's first cap for Wales, when under the captaincy of Frank Hill, the Welsh team beat the tourists five points to nil. Four days later, on 26 December, Harding led his Newport team against the Māori's, but without star player and Welsh legend Arthur Gould, Newport's supporters were not optimistic of success. They were proven right when the Māoris won three tries to nil.

In 1889, Harding was selected to represent Wales twice as part of the Home Nations Championship. Wales lost both games of the tournament against Scotland and Ireland, and Harding was not chosen to represent his country again.

International matches played

Wales (rugby union)

  1889
  \end{quote}
}
{
  \textbf{Paraphrased}\par
  \smallskip
  \begin{quote}
Charles Theodore "Theo" Harding (May 26, 1860 - July 13, 1919) was a renowned English-born rugby union player who showcased his skills at both club and international levels, representing Newport and Wales, respectively. Harding's versatility as an athlete extended beyond rugby, as he also led Newport Hockey Club to success in their inaugural season.

Rugby Career

As one of Newport's pioneering players, Harding assumed the role of captain for two consecutive seasons (1887/88 and 1888/89). During this period, he encountered the first-ever overseas touring team, the New Zealand Māoris, on two separate occasions. The initial encounter marked Harding's debut appearance for Wales, where, under the leadership of Frank Hill, the team emerged victorious with a score of five points to nil. Just four days later, on December 26, Harding led his Newport team against the Māoris, but the absence of the legendary Arthur Gould dampened the spirits of Newport's supporters, who correctly predicted a defeat, as the Māoris secured three tries to nil.

In 1889, Harding was selected to represent Wales in two Home Nations Championship matches. Unfortunately, Wales suffered losses against both Scotland and Ireland, and Harding did not receive further international call-ups.

International Matches Played

Wales (Rugby Union)

  1889
  \end{quote}
}
\end{tcolorbox}
\caption{Comparison of Original and Paraphrased Texts from the Wikipedia Dataset: An Example Featuring Charles Theodore `Theo' Harding's Rugby Career.}
\label{fig:wiki_example}
\end{figure*}


\begin{figure*}
\begin{tcolorbox}[title=Original and Paraphrased Text example for PubMed Central dataset, enhanced, raster equal height, boxrule=0.6pt, colback=white]
{
  \textbf{Original}\par
  \smallskip
  \begin{quote}
  Alzheimer's disease (AD), described for the first time by Alois Alzheimer in 1906, is characterized by progressive loss of cognitive functions ultimately leading to death[@b1]. Pathologically, the disease is characterized by the presence of extraneuronal amyloid plaques consisting of aggregates of amyloid-beta (A$\beta$) peptide, and neurofibrillary tangles (NFTs) which are intracellular aggregates of abnormally hyperphosphorylated tau protein[@b2]. A$\beta$ peptide is generated upon sequential cleavage of the amyloid precursor protein (APP), by beta- and gamma-secretases, and it is believed that an imbalance between A$\beta$ production and clearance results in its accumulation in the brain.

Clearance of A$\beta$ from the brain occurs via active transport at the blood-brain-barrier (BBB) and blood cerebrospinal fluid (CSF) barrier (BCSFB), in addition to the peptidolytic removal of the peptide by several enzymes. The receptors for A$\beta$ at the BBB bind A$\beta$ directly, or bind to one of its carrier proteins, and transport it across the endothelial cell. The low-density lipoprotein receptor-related protein 1 (LRP1) and the receptor for advanced glycation end products (RAGE) are involved in receptor-mediated flux of A$\beta$ across the BBB[@b3]. Both LRP1 and RAGE are multi-ligand cell surface receptors that, in addition to A$\beta$, mediate the clearance of a large number of proteins.
  \end{quote}
}
{
  \textbf{Paraphrased}\par
  \smallskip
  \begin{quote}
Alzheimer's disease, first identified by Alois Alzheimer in 1906, is marked by a gradual decline in cognitive abilities, ultimately resulting in mortality[@b1]. From a pathological perspective, the disease is distinguished by the presence of extraneuronal amyloid plaques comprising aggregated amyloid-beta peptides, as well as neurofibrillary tangles, which are intracellular accumulations of aberrantly hyperphosphorylated tau protein[@b2]. The amyloid-beta peptide is generated through the sequential cleavage of the amyloid precursor protein by beta- and gamma-secretases, and it is hypothesized that an imbalance between its production and clearance leads to its accumulation in the brain.

The removal of amyloid-beta from the brain occurs through active transport at the blood-brain barrier and blood-cerebrospinal fluid barrier, in addition to enzymatic degradation by various peptidases. At the blood-brain barrier, receptors for amyloid-beta bind the peptide directly or interact with one of its carrier proteins, facilitating its transport across the endothelial cell. The low-density lipoprotein receptor-related protein 1 and the receptor for advanced glycation end products play a role in receptor-mediated amyloid-beta flux across the blood-brain barrier[@b3]. Both of these receptors are multifunctional cell surface receptors that, in addition to amyloid-beta, mediate the clearance of numerous other proteins.
  \end{quote}
}
\end{tcolorbox}
\caption{Comparative Analysis of Original and Paraphrased Texts from the PubMed Central Dataset: Transthyretin participates in beta-amyloid transport from the brain to the liver- involvement of the low-density lipoprotein receptor-related protein 1?~\cite{alemi2016transthyretin}.}
\label{fig:pubmed_example}
\end{figure*}


\begin{figure*}
\begin{tcolorbox}[title=Original and Paraphrased Text example for PeS2o dataset, enhanced, raster equal height, boxrule=0.6pt, colback=white]
{
  \textbf{Original}\par
  \smallskip
  \begin{quote}
Molecular heterogeneity has great significance in the disease biology of multiple myeloma (MM). Thus, the analysis combined single-cell RNA-seq (scRNA-seq) and bulk RNA-seq data were performed to investigate the clonal evolution characteristics and to find novel prognostic targets in MM. The scRNA-seq data were analyzed by the Seurat pipeline and Monocle 2 to identify MM cell branches with different differentiation states. Marker genes in each branch were uploaded to the STRING database to construct the Protein-Protein Interaction (PPI) network, followed by the detection of hub genes by Cytoscape software. Using bulk RNA-seq data, Kaplan-Meier (K-M) survival analysis was then carried out to determine prognostic biomarkers in MM. A total of 342 marker genes in two branches with different differentiation states were identified, and the top 20 marker genes with the highest scores in the network calculated by the MCC algorithm were selected as hub genes in MM. Furthermore, K-M survival analysis revealed that higher NDUFB8, COX6C, NDUFA6, USMG5, and COX5B expression correlated closely with a worse prognosis in MM patients.
  \end{quote}
}
{
  \textbf{Paraphrased}\par
  \smallskip
  \begin{quote}
A comprehensive analysis integrating bulk RNA sequencing and single-cell RNA sequencing has shed light on novel prognostic biomarkers in multiple myeloma, a disease characterized by significant molecular heterogeneity. To investigate clonal evolution and identify potential targets, researchers combined single-cell RNA sequencing data, analyzed using the Seurat pipeline and Monocle 2, with bulk RNA sequencing data. This approach enabled the identification of distinct branches of multiple myeloma cells with varying differentiation states. The marker genes associated with each branch were used to construct a Protein-Protein Interaction network via the STRING database, and hub genes were detected using Cytoscape software. Subsequent Kaplan-Meier survival analysis, utilizing bulk RNA sequencing data, revealed prognostic biomarkers for multiple myeloma. A total of 342 marker genes were identified across two branches with differing differentiation states, and the top 20 genes with the highest scores, as calculated by the MCC algorithm, were selected as hub genes. Furthermore, survival analysis demonstrated that elevated expression of NDUFB8, COX6C, NDUFA6, USMG5, and COX5B correlated strongly with a poorer prognosis in multiple myeloma patients.  \end{quote}
}
\end{tcolorbox}
\caption{Original and Paraphrased Texts from the PeS2o Dataset: Integrative Analysis of Bulk RNA-Seq and Single-Cell RNA-Seq Unveils Novel Prognostic Biomarkers in Multiple Myeloma \cite{zhao2022integrative}}
\label{fig:pes2o_example}
\end{figure*}

\clearpage




\end{document}